\documentclass{article}

\usepackage{PRIMEarxiv}

\usepackage[utf8]{inputenc} 
\usepackage[T1]{fontenc}    
\usepackage{hyperref}       
\usepackage{url}            
\usepackage{booktabs}       
\usepackage{amsfonts}       
\usepackage{nicefrac}       
\usepackage{microtype}      
\usepackage{lipsum}
\usepackage{fancyhdr}       
\usepackage{graphicx}       
\usepackage[square,numbers,sort&compress]{natbib}

\graphicspath{{media/}}     

\title{Exploiting Diffusion Prior for out-of-distribution detection
}

\author{
  Armando Zhu \\
  Carnegie Mellon University \\
  USA\\
   \And
  Jiabei Liu \\
  North Eastern University \\
  USA\\
   \And
  Keqin Li \\
  AMA University \\
  Philippines\\
   \And
  Shuying Dai \\
  Indian Institute of Technology Guwahati \\
  India\\
   \And
  Bo Hong \\
  Northern Arizona University \\
  USA\\
   \And
  Peng Zhao \\
  Microsoft \\
  China\\
   \And
  Changsong Wei \\
  Digital Financial Information Technology Co.LTD \\
  China\\
}

\begin{document}
\maketitle

\begin{abstract}
Out-of-distribution (OOD) detection is crucial for deploying robust machine learning models, especially in areas where security is critical. However, traditional OOD detection methods often fail to capture complex data distributions from large scale date. In this paper, we present a novel approach for OOD detection that leverages the generative ability of diffusion models and the powerful feature extraction capabilities of CLIP. By using these features as conditional inputs to a diffusion model, we can reconstruct the images after encoding them with CLIP. The difference between the original and reconstructed images is used as a signal for OOD identification. The practicality and scalability of our method is increased by the fact that it does not require class-specific labeled ID data, as is the case with many other methods. Extensive experiments on several benchmark datasets demonstrates the robustness and effectiveness of our method, which have significantly improved the detection accuracy.
\end{abstract}

\keywords{Out of distribution detection \and CLIP \and Diffusion Model}

\section{Introduction}

The ability to identify out-of-distribution (OOD) data is a critical component in deploying robust machine learning models in real-world applications \cite{yan2024investigation, yang-24-data-aug, li2023stock, jmse11010007, wang2024advanced, liu2024rumor, xiang2024research, cheng2024research, zhang2024research}. OOD detection aims to identify instances that deviate significantly from the training distribution, ensuring the reliability of model predictions and minimizing the risk of erroneous outputs. This capability is particularly crucial in safety-critical domains such as autonomous driving \cite{ding-24-detection, unknown,zhou2023multimodal,zhou2023style,he2024lidar,mo2024make,dai2024cloud}, healthcare \cite{yuan2024research,chai2024deep,yao2024integrating,smucny2022deep,li2022automated,mo2024password}, and security systems, where the presence of unfamiliar data can lead to catastrophic failures.

Various approaches have been proposed to address the problem of OOD detection, ranging from statistical techniques to deep learning-based methods. Traditional methods often rely on simple feature extraction and anomaly detection algorithms, which may be inadequate for capturing complex data distributions.

Recently, the Contrastive Language–Image Pretraining (CLIP) model has emerged as a powerful backbone for feature extraction. CLIP leverages extensive internet data to learn rich, multimodal representations of images and text, demonstrating impressive zero-shot learning capabilities and effectiveness in various tasks without the need for task-specific fine-tuning.

Similarly, diffusion models have gained attention for their ability to generate high-quality images through a denoising process. These models learn the data distribution by progressively removing noise from a corrupted version of the image, effectively capturing the underlying data manifold. The prior knowledge embedded in diffusion models can be instrumental in reconstructing images and identifying anomalies. It can be beneficial for 3D vision tasks \cite{chen2023class, chen2021multimodal, chen2024bridging, chen2023point, Weng2024} and scene understanding \cite{xin2024vmt,deng2024compact,deng2023plgslam,shen2024localization,liu2024adaptive100,deng2023long}.

In this paper, we propose a novel approach to OOD detection by exploiting the diffusion prior. The core insight of our approach is that a model capable of accurately reconstructing an image indicates that the image is likely part of the distribution the model has learned. Conversely, poor reconstruction suggests that the image is out-of-distribution.

Our method involves utilizing the CLIP model to encode the image and using its features as conditional input for the diffusion model. By comparing the discrepancy between the reconstructed image and the original input, we can effectively determine if an image is OOD. This approach is based on the assumption that the model can only accurately reconstruct images of classes it has encountered during training, leveraging both the image input and its feature representation.

Additionally, for classification purposes, we utilize the zero-shot classification capability of CLIP, allowing us to classify images without fine-tuning the model. This is particularly advantageous as it enables the use of large amounts of in-distribution data without requiring labeled OOD data.

Our main contributions can be summarized as follows:

\begin{itemize}
\item We propose a novel OOD detection method based on the integration of CLIP and diffusion models.
\item We conduct extensive experiments on multiple benchmarks, demonstrating the robustness and efficacy of our method in OOD detection.
\end{itemize}

\section{Related Works}
\label{sec:headings}

\subsection{Out-of-Distribution Detection}

Several methods have been introduced to address the complex problem of OOD detection \cite{yang2021generalized, zhou2024optimizing, unknownPredict, shi2022deep, yang2024long, xu2023combination, ji2023prediction, ji2024rag}. A common strategy involves leveraging uncertainty estimation techniques, such as Bayesian modeling \cite{mackay1992practical}, to assess prediction uncertainty and identify OOD samples. Prominent techniques in this domain include Maximum Softmax Probabilities (MSP), which uses the maximum softmax output as a confidence measure \cite{hendrycks2016baseline}, Mahalanobis distance \cite{lee2018simple}, and Monte Carlo Markov Chain methods that facilitate sampling from high-dimensional distributions \cite{andrieu2003introduction}. Ensemble models are also widely acknowledged for enhancing the robustness and performance of machine learning systems, including OOD detection \cite{lakshminarayanan2017simple}. In OOD detection, ensemble methods integrate multiple base models for predictions, fitting into both probabilistic and uncertainty-based frameworks.

Supervised methods have shown some efficacy in reducing the incidence of erroneously high-confidence predictions on OOD inputs \cite{devries2018learning}; however, they are constrained by the necessity of labeled OOD data for training. Common unsupervised approaches include density estimation techniques \cite{kingma2018glow}. Recent research indicates that augmentation and adversarial perturbation can improve OOD detection performance \cite{choi2019novelty}. A key strength of our proposed OOD detection method is that it does not require specific class labels for training the diffusion model. Instead, it only requires in-distribution samples to learn the distribution, enabling it to determine whether a sample is in-distribution or OOD during testing.

\subsection{Pre-trained Vision-Language Models}

Interpreting the semantic information within images remains a significant challenge in computer vision \cite{xin2024parameter,xin2024mmap,xin2023self,deng2024incremental,zhang2024multi,gao2024enhanced,zhan2024innovations,yang2024new,li-19-segmentation,ding-24-style,ji2023improving}. The emergence of Transformers \cite{vaswani2017attention} has made a great impact on not only natural language processing field \cite{ji2024assertion, mei2024efficiency, xiao2024exploration, xu2024advancing, li-24-vqa, li-23-deception-detection, li2024feature,zhou2022towards,zhou2023thread,zhou2024fine,mo2024large,zhou2023improving,liu2024spam}, but also vision-related tasks \cite{dosovitskiy2020image}, paving the way for the introduction of CLIP \cite{radford2021learning}, a powerful pre-trained vision-language model. By utilizing contrastive learning along with extensive models and datasets \cite{schuhmann2022laion}, CLIP employs image-text pairs for self-supervised training. This strategy has effectively trained the model to align visual and textual representations within a latent space, facilitating robust feature extraction and zero-shot learning capabilities.

\subsection{Diffusion models}

Diffusion denoising probabilistic models, commonly known as diffusion models \cite{ho2020denoising}, have gained popularity as a notable class of generative models, recognized for their exceptional synthesis quality and controllability. The fundamental principle of these models involves training a denoising autoencoder to approximate the reverse of a Markovian diffusion process \cite{sohl2015deep}. By leveraging generative training on large-scale datasets with image-text pairs, such as LAION5B \cite{schuhmann2022laion}, diffusion models develop the ability to produce high-quality images featuring diverse content and coherent structures. Recently, a controllable architecture called ControlNet \cite{zhang2023adding} has been introduced, enabling the addition of spatial controls, such as depth maps and human poses, to pre-trained diffusion models, thereby expanding their applicability to controlled image generation.

\section{Method}
\label{sec:method}

\subsection{CLIP Model}

CLIP is a multi-modal vision and language model that has demonstrated impressive results in image-text similarity and zero-shot image classification, leveraging extensive training data and large-scale models. CLIP consists of an image encoder, such as CNN-based or Transformer-like models, and a causal language model to obtain text features. During the pre-training phase, CLIP uses large-scale image-text pairs for self-supervised contrastive learning, aligning images and texts into the same latent space.

In zero-shot image classification tasks, given $M$ class labels for classification (e.g., "cat", "dog"), CLIP incorporates these class labels into pre-designed hard/unlearnable text prompts, such as "a photo of a [class]", forming a prompt set like "a photo of a cat", "a photo of a dog", etc. These prompts are then fed into the text encoder to obtain $M$ text features $T_i$, where $i \in \{1, 2,... M\}$. The testing image is input into the image encoder to obtain an image feature $I_f$. The cosine similarity is calculated between the normalized image feature and all text features, formally, $Sim(T_i,I_f)=Tf \cdot If$, and the text feature $T_i$ with the highest similarity to $I_f$ is considered the image's category.

\begin{figure}
  \centering
  \includegraphics[width=0.65\textwidth]{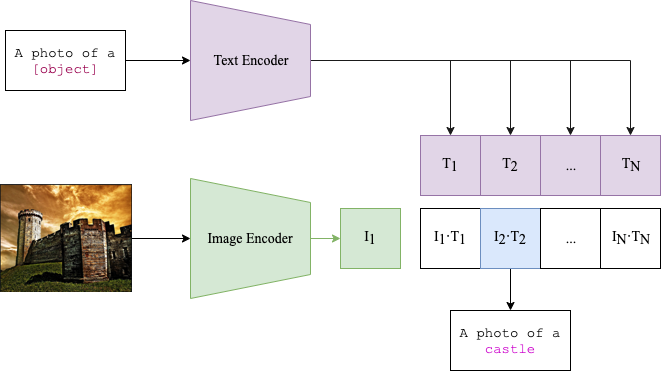}
  \caption{Architecture of CLIP model.}
  \label{fig:fig1}
\end{figure}

\subsection{Diffusion U-Net}

Diffusion Models are generative models used to generate data similar to the training data. Fundamentally, Diffusion Models work by progressively adding Gaussian noise to training data and then learning to recover the data by reversing this noising process.

Diffusion models achieve high controllability through effective cross-attention layers in the denoising U-Net, facilitating interactions between image features and various conditions. ControlNet, a neural network that enhances image generation in Stable Diffusion by adding extra conditions, allows users to control the images generated more precisely. ControlNet enhances the fine-grained spatial control on latent diffusion models (LDM) by leveraging a trainable copy of the encoding layers in the denoising U-Net as a strong backbone for learning diverse conditional controls.

During the training of the ControlNet framework, images are first projected to latent representations $z_0$ by a trained VQGAN consisting of the encoder (EEE) and the decoder (DDD). Denoting $z_s$ as the noisy image at the $s$-th timestep, it is produced by: 

\begin{equation}
z_s = \sqrt{\bar{a_t}}z_0 + \sqrt{1 - \bar{a_t}}\epsilon
\end{equation}

where $\bar{a_t} = \prod_{i=1}^{s}{a_i}$, and $\epsilon \sim N(0, I)$. By utilizing fine-grained conditions, ControlNet achieves controllable human image generation with various conditions based on the semantic information of the input.

\subsection{Proposed Out-of-Distribution Detection Method}

\begin{figure}
  \centering
  \includegraphics[width=0.9\textwidth]{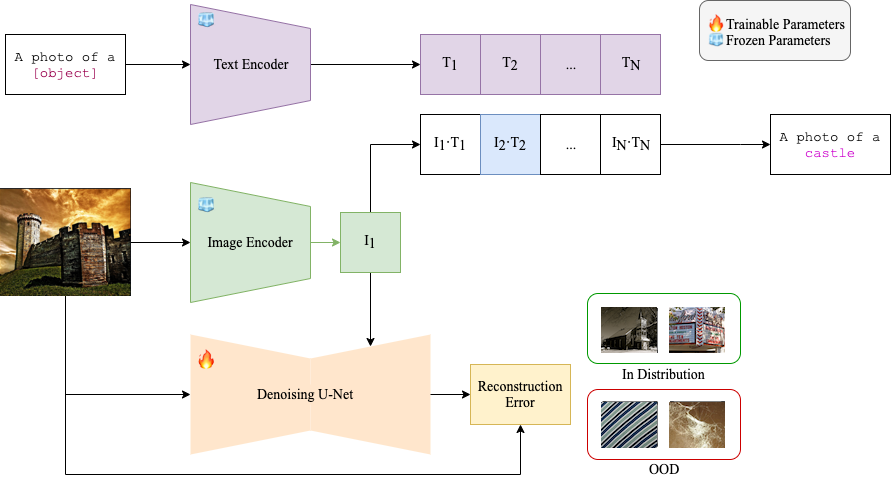}
  \caption{Architecture of the proposed method.}
  \label{fig:fig2}
\end{figure}

The features extracted from the CLIP model can be highly beneficial for classifying input images and distinguishing between in-distribution (ID) and out-of-distribution (OOD) samples.

We fine-tune a pre-trained denoising U-Net, guided effectively by the condition injection from features extracted by CLIP. The denoising U-Net is designed to reconstruct input images, and we use the reconstruction error to generate precision-recall curves for OOD detection. To prepare the input images, we convert the cropped image $I$ from pixel space to obtain the latent representation from the image encoder as part of the CLIP model. We then feed the image into the U-Net with guidance extracted from CLIP. The encoder takes grayscale input images of size $128 \times 128 \times 1$ and progressively reduces the spatial dimensions while increasing the number of channels, culminating in a bottleneck layer. The decoder then upscales and reconstructs the original input image through transposed convolutions and activations.

During training, the model is optimized to minimize the Mean Squared Error (MSE) loss between the reconstructed heatmaps and the original input. During inference, the threshold for distinguishing between in-distribution and out-of-distribution samples is set as the maximum reconstruction error of the in-distribution samples. With this approach, any sample with a reconstruction error above the threshold is classified as OOD, while samples below the threshold are considered in-distribution.

\section{Experiments}
\label{sec:Experiments}

\subsection{Experimental Details}

\subsubsection{Datasets}

To evaluate the efficacy of our proposed OOD detection method, we conducted extensive experiments using established benchmarks. We utilized the ImageNet-1K \cite{deng2009imagenet} dataset with 1,000 classes as the in-distribution (ID) dataset. For out-of-distribution (OOD) datasets, we selected subsets from Texture \cite{cimpoi2014describing}, iNaturalist \cite{van2018inaturalist}, Places \cite{zhou2017places}, and SUN \cite{xiao2010sun}, ensuring that the concepts in these datasets do not overlap with ImageNet-1K. Specifically, the entire Texture dataset was used for evaluation. Additionally, 110 plant classes not present in ImageNet-1K were selected from iNaturalist, 50 categories not present in ImageNet-1K were selected from Places, and 50 unique nature-related concepts were selected from SUN.

\subsubsection{Implementation Details}

We employed the CLIP model based on CLIP-B/16, pre-trained from OpenCLIP \cite{cherti2023reproducible}. For the denoising U-Net, we used Stable Diffusion V1-5 with ControlNet, pre-trained for image generation. The model was fine-tuned using ImageNet-1K samples for 10 epochs. During fine-tuning, the Mean Squared Error (MSE) loss was minimized between the reconstructed images and the original inputs to enhance the model's reconstruction capabilities.

\subsection{Comparison with Existing Models}

The results of OOD detection on the benchmark datasets are summarized in Table \ref{tab:performance_metrics}. Our proposed method consistently achieves superior or comparable performance across individual OOD datasets and in the averaged results. Compared with zero-shot methods, our approach surpasses the best competing method, CLIPN \cite{wang2023clipn}, by approximately 1.5\% in FPR95, despite CLIPN requiring an additional large external dataset to train an additional negative text encoder. Although our method is significantly more lightweight than CLIPN in model size, it consistently outperforms CLIPN in both metrics across all OOD datasets. Adapted post-hoc methods generally do not leverage CLIP's capabilities well and thus perform less effectively.

Our method also substantially surpasses prompt learning-based methods, reducing the FPR95 by about 23\%. This indicates that the learned diffusion model provides informed knowledge about OOD data, which is lacking in the competing methods, significantly reducing detection errors.

\begin{table*}[h!]
  \caption{Performance metrics across various datasets}
  \centering
  \resizebox{\textwidth}{!}{%
  \begin{tabular}{lcccccccccc}
    \toprule
    Dataset & \multicolumn{2}{c}{iNaturalist} & \multicolumn{2}{c}{SUN} & \multicolumn{2}{c}{Places} & \multicolumn{2}{c}{Texture} & \multicolumn{2}{c}{Average} \\
    \cmidrule(r){2-3} \cmidrule(r){4-5} \cmidrule(r){6-7} \cmidrule(r){8-9} \cmidrule(r){10-11}
    Metrics & FPR95 & AUROC & FPR95 & AUROC & FPR95 & AUROC & FPR95 & AUROC & FPR95 & AUROC \\
    \midrule
    \multicolumn{11}{l}{\textit{Zero-shot methods}} \\
    MCM \cite{ming2022delving} & 30.94 & 94.61 & 37.67 & 92.56 & 44.76 & 89.76 & 57.91 & 86.1 & 42.82 & 90.76 \\
    GL-MCM \cite{miyai2023zero} & 15.18 & \textbf{96.71} & 30.42 & 93.09 & 38.85 & 89.9 & 57.93 & 83.63 & 35.47 & 90.83 \\
    CLIPN \cite{wang2023clipn} & 23.94 & 95.27 & 26.17 & 93.92 & 33.45 & \textbf{92.28} & \textbf{40.83} & 90.93 & 31.1 & 93.1 \\
    \midrule
    \multicolumn{11}{l}{\textit{CLIP-based posthoc methods}} \\
    MSP \cite{hendrycks2016baseline} & 74.57 & 77.74 & 76.95 & 73.97 & 79.12 & 72.18 & 73.66 & 74.84 & 76.22 & 74.98 \\
    MaxLogit \cite{hendrycks2019scaling} & 60.88 & 88.03 & 44.83 & 91.16 & 55.54 & 87.45 & 48.72 & 88.63 & 52.49 & 88.82 \\
    ODIN \cite{liang2017enhancing} & 30.22 & 94.65 & 54.04 & 87.17 & 55.06 & 85.54 & 51.67 & 87.85 & 47.75 & 88.8 \\
    ViM \cite{wang2022vim} & 32.19 & 93.16 & 54.01 & 87.19 & 60.67 & 83.75 & 53.94 & 87.18 & 50.2 & 87.82 \\
    KNN \cite{sun2022out} & 29.17 & 94.52 & 35.62 & 92.67 & 39.61 & 91.02 & 64.35 & 85.67 & 42.19 & 90.97 \\
    \midrule
    \multicolumn{11}{l}{\textit{Prompt Learning Methods}} \\
    CoOp \cite{zhou2022conditional} & 29.81 & 93.77 & 40.83 & 93.29 & 40.11 & 90.58 & 45 & 89.47 & 51.68 & 91.78 \\
    \midrule
    Ours & \textbf{15.03} & 96.45 & \textbf{24.95} & \textbf{94.59} & \textbf{33.17} & 90.83 & 41.85 & \textbf{91.02} & \textbf{28.75} & \textbf{93.25} \\
    \bottomrule
  \end{tabular}}
  \label{tab:performance_metrics}
\end{table*}

\section{Conclusion}

In this paper, we introduced a novel approach to out-of-distribution (OOD) detection by combining the feature extraction capabilities of CLIP with the generative power of diffusion models. Our method involves encoding images with CLIP and using these features as conditional inputs for a diffusion model to reconstruct the images. The discrepancy between the original and reconstructed images serves as a robust indicator for OOD detection.

Our approach offers several advantages over existing methods. Firstly, it does not require labeled OOD data, making it more practical and scalable for real-world applications. By leveraging only in-distribution samples for training, our method effectively discerns between in-distribution and OOD samples during testing. Secondly, the integration of CLIP's zero-shot classification capability enhances the versatility of our method, allowing for effective image classification without the need for model fine-tuning.

We conducted extensive experiments on multiple benchmark datasets, including ImageNet-1K, Texture, iNaturalist, Places, and SUN. The results demonstrate that our method achieves significant improvements in detection accuracy, with substantial reductions in false positive rates and enhanced detection metrics across diverse datasets. These findings underscore the potential of integrating pre-trained models to enhance the reliability of OOD detection, paving the way for the deployment of more dependable machine learning systems.

Future work could explore further enhancements to our method, such as incorporating additional types of pre-trained models or refining the reconstruction process to improve detection accuracy further. Additionally, applying our approach to other domains beyond image data, such as natural language processing or audio data, could provide valuable insights and broaden the applicability of our technique.



\bibliographystyle{unsrtnat}  
\bibliography{references.bib}  


\end{document}